\documentclass{article}

\usepackage[preprint]{neurips_2025}

\usepackage[utf8]{inputenc} 
\usepackage[T1]{fontenc}    

\usepackage{url}            
\usepackage{booktabs}       
\usepackage{amsfonts}       
\usepackage{nicefrac}       
\usepackage{microtype}      
\usepackage[pagebackref,breaklinks,colorlinks]{hyperref}
\usepackage{multirow,bbding}
\usepackage[table,xcdraw]{xcolor}
\usepackage{amsmath}
\usepackage{makecell}
\usepackage{longtable}
\usepackage{graphicx}
\usepackage{wrapfig}
\usepackage{algorithm}
\usepackage{algorithmicx}
\usepackage{algpseudocode}
\usepackage{soul}

\newcommand{\ourmodel}{\textit{V-Reflection}}
\newcommand{\lvrbox}[1]{%
    \definecolor{graybg}{gray}{0.9}%
    \sethlcolor{graybg}%
    \hl{\texttt{#1}}%
}
\title{V-Reflection: Transforming MLLMs from Passive Observers to Active Interrogators}

\author{
  \textbf{Jiazhou Zhou}$^{1,2}$\thanks{Work done during an internship at IDEA Research.} \quad
  \textbf{Yucheng Chen}$^{3}$ \quad
  \textbf{Hongyang Li}$^{4,2}$\textsuperscript{*} \quad
  \textbf{Qing Jiang}$^{4,2}$\textsuperscript{*} \\
  \textbf{Hu Zhou}$^{5}$ \quad
  \textbf{Ying-Cong Chen}$^{1}$ \quad
  \textbf{Lei Zhang}$^{2}$\thanks{Corresponding Author.} \\
  \\
  $^1$AI Thrust, The Hong Kong University of Science and Technology (Guangzhou) \\
  $^2$International Digital Economy Academy \\
  $^3$MedVisAI Lab, Lee Kong Chian School of Medicine, \\ Nanyang Technological University, and Centre of AI in Medicine \\
  $^4$South China University of Technology \\
  $^5$Department of Electrical and Electronic Engineering, \\
  The Hong Kong Polytechnic University \\
  \\
  \normalsize Project page: \url{https://idea-research.github.io/V-Reflection/}
}

\begin{document}
\maketitle

\vspace{-15pt}
\begin{abstract}
Multimodal Large Language Models (MLLMs) have achieved remarkable success, yet they remain prone to perception-related hallucinations in fine-grained tasks. This vulnerability arises from a fundamental limitation: their reasoning is largely restricted to the language domain, treating visual input as a static, reasoning-agnostic preamble rather than a dynamic participant. Consequently, current models act as passive observers, unable to re-examine visual details to ground their evolving reasoning states. To overcome this, we propose \ourmodel{}, a framework that transforms the MLLM into an active interrogator through a "think-then-look" visual reflection mechanism. During reasoning, latent states function as dynamic probes that actively interrogate the visual feature space, grounding each reasoning step for task-critical evidence. Our approach employs a two-stage distillation strategy. First, the Box-Guided Compression Module (BCM) establishes stable pixel-to-latent targets through explicit spatial grounding. Next, a Dynamic Autoregressive Compression (DAC) module maps the model's hidden states into dynamic probes that interrogate the global visual feature map. By distilling the spatial expertise of the BCM teacher into the DAC student, \ourmodel{} internalizes the ability to localize task-critical evidence. During inference, both modules remain entirely inactive, maintaining a purely end-to-end autoregressive decoding in the latent space with optimal efficiency. Extensive experiments demonstrate the effectiveness of our \ourmodel{} across six perception-intensive benchmarks, significantly narrowing the fine-grained perception gap. Visualizations confirm that latent reasoning autonomously localizes task-critical visual evidence.
\end{abstract}

\section{Introduction}
\label{sec:intro}
Multimodal Large Language Models (MLLMs) \cite{bai2023qwen,bai2025qwen3,chen2024internvl,zhu2025internvl3,wang2025internvl3,li2024llava,hurst2024gpt} have achieved remarkable success in bridging intelligence with visual understanding, demonstrating sophisticated capabilities in cross-modal alignment and complex instruction following \cite{fu2024blink,xu2025mc,wu2024v,zhang2024mme,hrbench,tong2024eyes}. Despite these advancements, current MLLMs remain prone to perception-related hallucinations in fine-grained tasks. This limitation is primarily rooted in a language-centric reasoning paradigm: the prevailing LLaVA-style architecture \cite{li2024llava} utilizes the pre-trained vision encoder that compresses complex visual inputs into static features, and once the autoregressive reasoning starts, the model loses the ability to retrace fine-grained details discarded during the initial encoding stage. 

\begin{figure*}[tb]
  \centering
  \includegraphics[height=3.6cm]{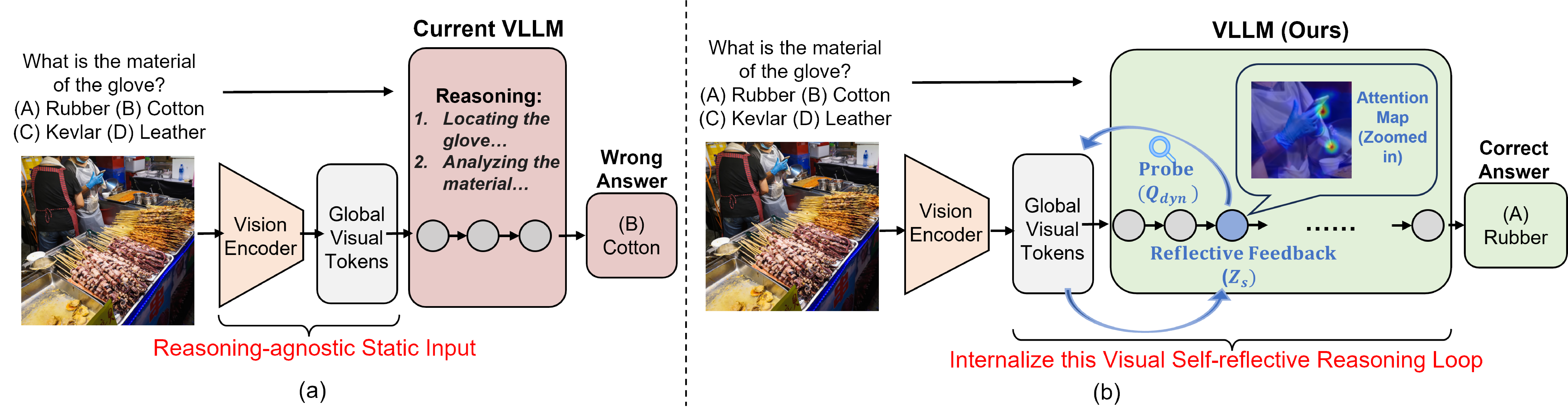}
\vspace{-10pt}
  \caption{Conceptual comparison between traditional MLLMs and our \ourmodel{}. Current MLLMs' reasoning remains confined to the language domain, treating visual information as a reasoning-agnostic static input rather than an active driver of the thought process, thus leading to perception-related hallucinations (e.g., `Kevlar') where the model prioritizes language priors over actual visual evidence. (b) Our framework internalizes a "think-then-look" visual self-reflection mechanism, where evolving latent states act as dynamic probes ($\mathbf{Q}_{dyn}$) to retrace global visual features. This mechanism retrieves task-specific evidence (e.g., accurately localizing the rubber glove), effectively correcting the reasoning trajectory for a precise answer.}
  \label{fig:teaser}
    \vspace{-15pt}
\end{figure*}

As illustrated in Fig. \ref{fig:teaser} (a), for instance, the current MLLM may misidentify a texture as "cotton" due to its high-frequency co-occurrence with "glove" in the pretraining corpus, even when the visual evidence provides a clear indication for "rubber." In this way, the MLLMs act as passive observers: they treat visual information as a reasoning-agnostic preamble rather than leveraging it as a dynamic resource to guide the reasoning trajectory.

Existing multimodal reasoning paradigms generally fall into two categories to address this limitation. "Thinking about Images" conducts reasoning in the language space, evolving from SFT-based patterns \cite{xu2025llava, shao2024visual} to RL-optimized trajectories \cite{deng2025emerging, yang2025r1, peng2025lmm, tan2025reason} and visual grounding by predicting points, bounding boxes, or descriptions \cite{yu2025perception, liu2025visual, jiang2025rex, ni2025point}. Meanwhile, "Thinking with Images" employs external tool APIs (e.g., OCR, cropping, programs) \cite{zhang2025chain, wang2025pixel, zheng2025deepeyes, su2025openthinkimg, wu2025vtool}, which are limited to tool availability and poor generalizability. Crucially, both paradigms treat visual information as a static input rather than an active partner; as a result, neither fundamentally resolves perception-related hallucinations.

Inspired by recent advances in latent reasoning \cite{deng2025emerging, huang2025vision}, which utilize continuous hidden manifolds to refine reasoning trajectories, we identify a promising path to overcome current bottlenecks. We introduce a "think-then-look" visual reflection mechanism. Within this framework, latent states function as dynamic probes that actively interrogate the visual feature space, grounding each step of the reasoning process for task-critical evidence. To this end, we propose \ourmodel{}, a framework that transforms the MLLM into an active interrogator. As illustrated in Fig. \ref{fig:teaser} (b), \ourmodel{} localizes task-critical features (e.g., the specific texture of the glove) and derives a precise answer.

Our approach employs a two-stage distillation strategy to synthesize explicit visual grounding with latent reasoning. In Stage 1 (Explicit Grounding Warm-up), the primary goal is to establish a stable pixel-to-latent for a high-quality supervision target. As illustrated in Fig. \ref{fig:Architecture} (a), we introduce the Box-Guided Compression Module (BCM). The BCM utilizes RoI-Align to extract local features $\mathbf{F}_{local}$ based on bounding boxes $\mathcal{B}$, which are then compressed into teacher latent tokens $\mathbf{Z}_T$ via cross-attention. In this way, the BCM compresses redundant visual patches into a structured set of latent tokens, serving as the foundation for aligning the MLLM’s continuous reasoning trajectory within the latent space.

While the BCM provides precise grounding, it is constrained by the requirement of bounding box priors. To enable the model to autonomously localize task-related evidence within global scenes, we transition to Stage 2 (Visual Latent Distillation). Here, we introduce the Dynamic Autoregressive Compression (DAC) Module. As illustrated in Fig. \ref{fig:Architecture} (b), the DAC projects hidden states $\mathbf{H}$ into dynamic probes $\mathbf{Q}_{dyn}$ that interrogate the global feature $\mathbf{F}_{global}$ through cross-attention to generate the student latent token $Z_{s}$. We then distill the BCM teacher into DAC student via minimizing two objectives: (1) The Mean Squared Error (MSE) between $\mathbf{Z}_S$ and $\mathbf{Z}_T$ to ensure representational alignment; (2) The KL-divergence between the student’s global attention maps and the teacher’s localized attention maps to transfer spatial priors.

This two-stage distillation paradigm allows \ourmodel{} to fully internalize spatial grounding expertise within its latent reasoning process. Crucially, both the BCM and DAC modules remain inactive during inference. As detailed in Tab. \ref{tab:efficiency}, this end-to-end reasoning paradigm introduces no extra architectures, ensuring the visual self-reflection process is executed with optimal inference efficiency.

Extensive experiments demonstrate the effectiveness of \ourmodel{} across six perception-intensive benchmarks. As shown in Tab. \ref{table:bench_table1}, \ourmodel{} achieves 72.3\% on MMVP, showing a clear margin over GPT-4o (58.33\%) and baseline Qwen2.5-VL (66.7\%). It also shows strong adaptability in high-resolution scenarios, with a +6.7\% increase on the HRBench-4K (FCP) subset Tab. \ref{table:bench_table2}, which effectively narrows fine-grained perception gaps. Furthermore, the visualization results in Fig. \ref{fig4-inference-atten} provide qualitative evidence of visual self-reflection, where the model autonomously focuses on task-related visual evidence driven by its internal reasoning process rather than Bbox priors.

In summary, our key contributions are as follows:

\noindent \textbf{(1)} We propose \ourmodel{}, a framework that transforms MLLMs into active interrogators via a visual self-reflection mechanism, where the model's latent states drive its visual focus to retrieve task-specific evidence during the latent reasoning process.

\noindent \textbf{(2)} We introduce a two-stage distillation strategy to synthesize explicit visual grounding with continuous latent reasoning. The (BCM) module first establishes stable pixel-to-latent targets through explicit spatial grounding. Then the DAC distills the spatial expertise of the BCM teacher to internalize the ability to autonomously localize task-critical evidence by latent states. During inference, both modules remain entirely inactive, maintaining a purely end-to-end autoregressive decoding with optimal efficiency.

\noindent \textbf{(3)} Extensive experiments prove the efficacy of \ourmodel{} across six perception-intensive benchmarks. Visualizations further confirm its ability to autonomously pinpoint task-critical pixels driven by the latent reasoning process.

\section{Related Work}
\label{sec:related}

\subsection{Multimodal Chain-of-Thought}
Chain-of-Thought (CoT) reasoning has evolved from text-only paradigms \cite{wei2022chain} to complex multimodal contexts \cite{shao2024visual}. We categorize these developments into two primary lines of research based on their interaction with visual information: \textbf{(1) Think about Images.} This line of work performs reasoning primarily in the language space. Early approaches focused on supervised fine-tuning (SFT) to acquire reasoning patterns \cite{xu2025llava, shao2024visual}. More recently, the field has shifted toward RL-based methods \cite{deng2025emerging, yang2025r1,peng2025lmm,tan2025reason,yang2025r1} to optimize the reasoning trajectory via textual proxies. Some studies emphasize visual grounding by predicting points, bounding boxes, or descriptions \cite{yu2025perception,liu2025visual,jiang2025rex,ni2025point} to ensure the model focuses on regions of interest (ROIs). While effective, "thinking about images" remains an indirect and inefficient representation, as it compels the model to translate rich visual evidence into discrete text before reasoning. \textbf{(2) Think with Images.} To overcome the limitations of text-only reasoning, other research approach augment MLLMs with predefined visual tools. These approaches employ utilities such as cropping, zooming, OCR engines, or chart parsers, \cite{zhang2025chain,wang2025pixel,zhang2025chain} to retrieve fine-grained details. Recent efforts utilize reinforcement learning to decide when to invoke these external APIs \cite{zheng2025deepeyes,su2025openthinkimg,su2025openthinkimg,wu2025vtool}, enabling interleaved CoT reasoning and tool execution. However, these methods are fundamentally constrained by the availability and design of external tools; tool APIs are often difficult to extend and require substantial training effort to adapt to new tasks. In summary, both categories treat visual information as a static input rather than an active driver of the reasoning trajectory. This language-centered bottleneck prevents the model from dynamically re-interrogating the visual scene, leaving a critical gap between fixed perceptual features and the fluid, evolving needs of the reasoning process.

\subsection{Latent Reasoning in MLLMs}
Recently, a shift in reasoning paradigms for LLM has moved beyond discrete token sequences toward continuous latent streams \cite{hao2024training, shen2025codi}, enabling models to navigate high-dimensional manifolds for greater flexibility and more condensed reasoning chains \cite{cheng2024compressed}. Building on this direction, several works have extended latent reasoning to MLLMs. Current methods primarily align these states with static encoder features or auxiliary signals, such as helper images \cite{yang2025machine, wang2025monet}, annotated boxes \cite{li2025latent} or fine-grained perceptual priors from models \cite{qin2025chain}. However, these introduced supervision signals necessitate costly auxiliary data and critically overlook the dynamic, top-down guidance inherent in the LLM’s own evolving latent states. In contrast, \ourmodel{} replaces such passive alignment with an active interrogation paradigm, utilizing hidden states as dynamic probes to autonomously scrutinize the visual space for a reasoning-aware search that eliminates auxiliary dependencies.

\vspace{-10pt}
\section{Methodology}

\subsection{Preliminary \& Overview}
\noindent\textbf{Preliminary for Coconut.} 
Coconut \cite{hao2024training} reformulates Chain-of-Thought (CoT) by shifting from discrete tokens to a continuous latent thought stream. Specifically, within a segment encapsulated by \lvrbox{<bot>} and \lvrbox{<eot>} tokens, the LLM autoregressively generates latent states $\mathbf{h}_i \in \mathbb{R}^D$ that are refed directly as input embeddings for subsequent steps without decoding into a vocabulary token via the LM head. This recursive feedback allows the model to utilize a high-dimensional latent space for advanced reasoning patterns such as breadth-first search (BFS). This capability is acquired via a progressive training strategy: in each Stage $k$, the first $k$ tokens of CoT are replaced by $k$ continuous states, forcing the model to gradually distill complex semantic planning and reasoning CoT into the latent CoT.

\begin{figure*}[t]
\centering
\includegraphics[height=8.5cm]{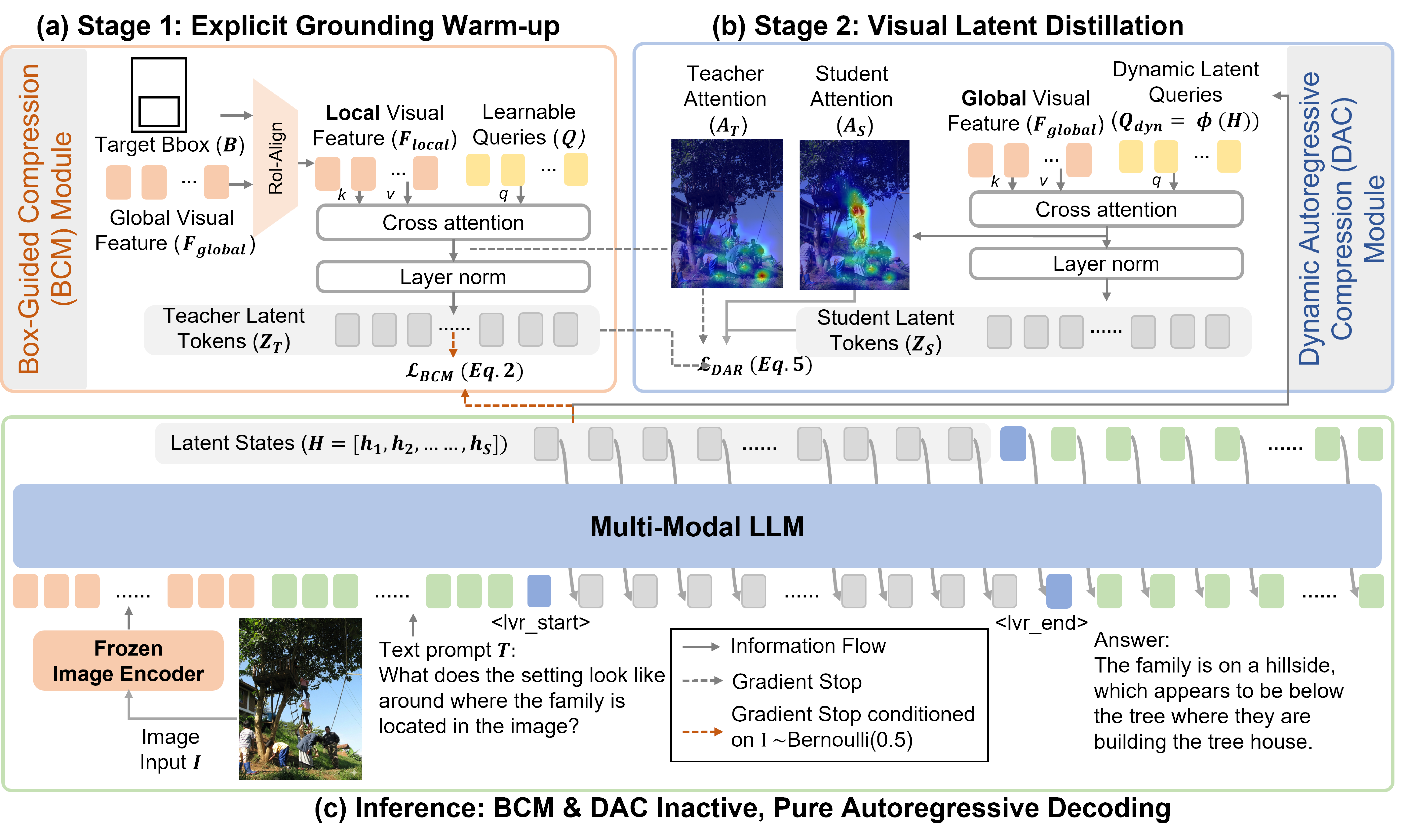}
\vspace{-15pt}
\caption{The V-Reflection Architecture. A two-stage paradigm establishes a "think-then-look" visual self-reflection reasoning mechanism. \textbf{(a)} Stage 1: The Box-Guided Compression (BCM) module distills regional patches into grounded latent tokens $\mathbf{Z}_T$ via $\mathcal{L}_{BCM}$. \textbf{(b)} Stage 2: The Dynamic Autoregressive Compression (DAC) module distills the spatial expertise of the BCM module, training the LLM's hidden states $\mathbf{H}$ to act as dynamic probes that autonomously interrogate global features. \textbf{(c)} Inference: Both BCM and DAC remain entirely inactive, as they have been fully internalized to execute a purely end-to-end visual search driven by its latent reasoning process.}
\label{fig:Architecture}
\end{figure*}

\noindent\textbf{Overview of \ourmodel{}.} In this paper, we extend Coconut \cite{hao2024training} to the multimodal domain. We propose V-Reflection, a framework designed to bridge implicit latent
reasoning and explicit grounding. We introduce the Box-Guided Compression Module (BCM) to distill redundant visual features into grounded, compressed latent tokens (Sec. \ref{sec:3.2}). Furthermore, the Dynamic Autoregressive Compression (DAC) is proposed to bridge explicit grounding and implicit reasoning (Sec. \ref{sec:3.3}). Through a multi-stage training paradigm (Sec. \ref{sec:3.4}), we realize the visual self-reflection mechanism, enabling the LLM to autonomously trigger visual probes within the latent reasoning process. The following sections will elaborate on these parts.

\subsection{Grounding Latent tokens via Box-Guided Compression}
\label{sec:3.2}
To bridge implicit latent reasoning with explicit grounding, we first introduce the Box-Guided Compression Module (BCM). Functioning as a high-fidelity teacher, the BCM establishes stable pixel-to-latent alignment by condensing redundant visual patches into a set of latent features. This process creates a stable target for aligning the LLM’s reasoning trajectory within the continuous latent space.

\noindent \textbf{Module Architecture.}
As shown in Fig.~\ref{fig:Architecture}, given an input image $I$ and a text prompt $T$, the LLM processes the multimodal sequence autoregressively. Following the Coconut \cite{hao2024training}, we define a specialized latent visual reasoning segment delimited by \lvrbox{lvr\_start} and \lvrbox{lvr\_end} tokens. Specifically, the \lvrbox{lvr\_start} token triggers the transition from discrete language generation to latent reasoning. Within this reasoning process, at each reasoning step $i$, the MLLM generates a latent state $\mathbf{h}_i \in \mathbb{R}^{1 \times D}$. During the first training phase, as present in Fig.~\ref{fig:Architecture} (a), to provide explicit spatial guidance, we utilize the target bounding box $\mathcal{B}$ as a regional prior. From the global visual feature map $\mathbf{F}_{global} \in \mathbb{R}^{L \times D}$, local region features $\mathbf{F}_{local} \in \mathbb{R}^{N \times D}$ are extracted via RoI-Align. To compress the image features and filter noise and redundancy, the BCM employs $S$ learnable queries $\mathbf{Q} \in \mathbb{R}^{S \times D}$, which function as learnable perceptual probes. These probes compress the local region features $\mathbf{F}_{local}$ into the teacher latent tokens $\mathbf{Z}_T \in \mathbb{R}^{S \times D}$:
\begin{equation}
\mathbf{Z}_{T} = \text{LayerNorm}(\text{CrossAttn}(\mathbf{Q}, \mathbf{F}_{local}, \mathbf{F}_{local})).
\end{equation}
Through standard cross attention, this architecture transforms unstructured visual signals into compressed latent tokens. Upon reaching the \lvrbox{lvr\_end} token, the model terminates the latent visual reasoning process and switches to normal discrete decoding to produce the final answer.

\noindent \textbf{Stochastic Decoupled Alignment Strategy.} To bridge the gap between explicit visual grounding and implicit reasoning trajectories, we enforce mutual alignment between $\mathbf{Z}_T$ and $\mathbf{H}$. However, a naive joint optimization of both $\mathbf{Z}_T$ and $\mathbf{H}$ frequently triggers representation collapse. This phenomenon occurs because, in a symmetric loss function without constraints, the two independent latent spaces tend to find a shortcut by converging to a trivial, non-informative constant. 

To ensure optimization stability, we propose a stochastic decoupled alignment strategy. Instead of updating both modules simultaneously, we treat $\mathbf{Z}_T$ and $\mathbf{H}$ as alternating optimization targets, effectively "freezing" one while the other adapts. To implement this efficiently within a single training pipeline, we introduce a binary indicator variable $\mathbb{I} \sim \text{Bernoulli}(0.5)$ to stochastically switch the gradient flow in each iteration:
\begin{equation}
\mathcal{L}_{BCM} = \mathbb{I} \cdot |\text{sg}(\mathbf{Z}_T) - \mathbf{H}|_1 + (1 - \mathbb{I}) \cdot |\mathbf{Z}_T - \text{sg}(\mathbf{H})|_1.
\end{equation}
By utilizing the stop-gradient operator $\text{sg}(\cdot)$ conditioned on $\mathbb{I}$, we ensure that in any given step, only one target is optimized. This decoupled strategy prevents the collapse into local minima, forcing the BCM to produce spatially grounded features while the LLM learns to project its reasoning trajectory into the corresponding latent space. This establishes the high-quality supervision required for subsequent pixel-level distillation.

\subsection{Visual Latent Distillation via Dynamic Autoregressive Compression}
\label{sec:3.3}
While the BCM establishes a high-fidelity supervision target for latent alignment, it is constrained by the bounding-box prior. This dependency on external boxes precludes the model from autonomously localizing task-relevant evidence within a global visual context without human-annotated priors. 

To address this, we introduce the Dynamic Autoregressive Compression (DAC) module, which distills the spatial grounding expertise of the BCM module. By leveraging the LLM’s evolving latent states $\mathbf{h}_i$ as dynamic probes to interrogate the global visual feature map $\mathbf{F}_{global}$, DAC enables the model to get rid of grounding box priors. This transforms the LLM from a passive observer into an active interrogator, where the reasoning trajectory—rather than external coordinates—autonomously drives visual focus to resolve logical ambiguities.

\noindent \textbf{Dynamic Latent Queries.}
In contrast to the BCM’s use of static learnable queries to compress local features, the Dynamic Autoregressive Compression (DAC) module leverages the LLM's internal hidden states to actively steer its focus across the global visual context. This design enables task-related visual focusing, where the model’s attention is adaptively modulated by the specific reasoning requirements of its current thought process. 

Specifically, during the latent reasoning trajectory, we collect $S$ steps of latent states $\mathbf{H} = [\mathbf{h}_1, \dots, \mathbf{h}_S] \in \mathbb{R}^{S \times D}$, which represent a continuous representation of the model’s evolving visual information needs. We project these states into a dynamic query space $\mathbf{Q}_{dyn} = \phi(\mathbf{H})$, where $\phi$ is a learnable linear projection. The resulting student latent tokens $\mathbf{Z}_S \in \mathbb{R}^{S \times D}$ are then computed via cross-attention over the global visual feature map $\mathbf{F}_{global} \in \mathbb{R}^{L \times D}$:
\begin{equation}
\mathbf{Z}_S = \text{LayerNorm}(\text{CrossAttn}(\mathbf{Q}_{dyn}, \mathbf{F}_{global}, \mathbf{F}_{global})).
\end{equation}
By employing each latent state $\mathbf{h}_i$ as a targeted probe, the model autonomously retrieves refined visual evidence that is strictly aligned with its current reasoning intent, effectively closing the loop between thinking and seeing.

\noindent \textbf{Visual Latent Distillation.}
To internalize spatial grounding without requiring explicit bounding boxes during inference, we distill the expertise of the BCM teacher into the DAC student via two complementary objectives: spatial prior transfer and representational alignment.

Before aligning the spatial focus, we define the attention mechanisms for both the teacher and student modules. For the BCM teacher, the localized attention $\mathcal{A}_T$ is computed by querying the local features $\mathbf{F}_{local}$ with the learnable queries $\mathbf{Q}$, namely $\mathcal{A}_T = \text{Softmax}\left(\frac{\mathbf{Q} \mathbf{K}_{local}^\top}{\sqrt{d}}\right)$. Conversely, for the DAC student, the global attention $\mathcal{A}_S$ is derived by using the dynamic probes $\mathbf{Q}_{dyn}$ to interrogate the global feature map $\mathbf{F}_{global}$, namely $\quad \mathcal{A}_S = \text{Softmax}\left(\frac{\mathbf{Q}_{dyn} \mathbf{K}_{global}^\top}{ \sqrt{d}}\right)$. $\mathbf{K}_{local}$ and $\mathbf{K}_{global}$ are the key projections of the localized and global visual features, respectively, and $\tau$ is a temperature hyperparameter used to soften the student's distribution.

To resolve the dimensionality mismatch between the teacher’s localized attention $\mathcal{A}_T \in \mathbb{R}^{S \times N}$ and the student’s global attention $\mathcal{A}_S \in \mathbb{R}^{S \times L}$, we construct a global target distribution $\hat{\mathcal{A}}_T \in \mathbb{R}^{S \times L}$ by projecting the teacher’s grounded focus onto the global spatial grid. Specifically, for each query $i$, the target weight $\hat{a}_{T, (i, j)}$ is assigned the value $a_{T, (i, k)}$ if the $j$-th global patch spatially coincides with the $k$-th patch within the grounding box $\mathcal{B}$; otherwise, it is set to zero. 

We then apply a temperature-scaled Softmax over $\mathcal{A}_S$ and employ Kullback-Leibler (KL) divergence to penalize the student’s deviation from the teacher's grounded focus for spatial prior transfer:
\begin{equation}\mathcal{L}_{attn} = \sum_{i=1}^{S} D_{KL}(\hat{\mathcal{A}}_{T, i} \parallel \mathcal{A}_{S, i}) = \sum_{i=1}^{S} \sum_{j=1}^{L} \hat{a}_{T, (i, j)} \log \left( \frac{\hat{a}_{T, (i, j)} + \epsilon}{a_{S, (i, j)} + \epsilon} \right),
\end{equation}
where $\epsilon$ is a smoothing constant to ensure numerical stability. By minimizing $\mathcal{L}_{attn}$, DAC is forced to suppress irrelevant background noise and concentrate on task-relevant regions.

Parallel to attention distillation, we enforce an $L_1$ loss between the student tokens $\mathbf{Z}_S$ and the teacher tokens $\mathbf{Z}_T$ to ensure that the compressed features are semantically consistent with the grounded targets. The total objective for the DAC module is summarized as follows:
\begin{equation}
\mathcal{L}_{\text{DAC}} = \lambda_{\text{feat}} \cdot |\mathbf{Z}_S - \text{sg}(\mathbf{Z}_T)|_1 + \lambda_{\text{attn}} \cdot \mathcal{L}_{\text{attn}},
\end{equation}
where $\lambda_{\text{feat}}$ and $\lambda_{\text{attn}}$ are hyperparameters that balance the trade-off between semantic alignment and spatial grounding; $\text{sg}(.)$ denotes the stop gradient operator. This formulation ensures that DAC inherits the BCM's precision while gaining the flexibility to operate on global features.

\subsection{Intriguing Visual Self-Reflection: A Two-Stage Supervision}
\label{sec:3.4}
To synergize the foundational grounding of the BCM with the adaptive flexibility of the DAC, we propose a two-stage training paradigm. This strategy systematically transitions the model from explicit, box-constrained perception to autonomous latent visual reasoning, ensuring that the DAC inherits the teacher’s spatial precision while gaining the freedom to operate on global visual features.

\begin{itemize}
\item \textbf{Stage 1: Explicit Grounding Warm-up.} The BCM and the LLM backbone are activated in this stage. Guided by ground-truth bounding boxes $\mathcal{B}$, the BCM module optimizes $\mathcal{L}_{stage1} = \mathcal{L}_{CE} + \lambda_{BCM} \cdot \mathcal{L}_{BCM}$ (Sec. \ref{sec:3.2}). where $\mathcal{L}_{CE}$ is the standard cross-entropy loss. This stage establishes a shared representational space where teacher latent tokens $\mathbf{Z}_T$ and latent states $\mathbf{H}$ are mutually aligned.

\item \textbf{Stage 2: Visual Latent Distillation.} We freeze the BCM teacher and keep the DAC student and the LLM backbone activated. The model is trained on the same dataset as stage 1 without bounding boxes to minimize: $\mathcal{L}_{stage2} = \mathcal{L}_{CE} + \lambda_{DAC} \cdot \mathcal{L}_{DAC}$ (Sec. \ref{sec:3.3}). By distilling both semantic features and pixel-level attention maps, DAC learns to mimic the BCM’s localized search directly on $\mathbf{F}_{global}$, internalizing the ability to resolve logical ambiguities.
\end{itemize}

\noindent \textbf{Inference: Visual Self-Reflection Reasoning.}
Our framework realizes the "think-then-look" self-reflection mechanism with efficiency. During inference, both the BCM and DAC distillation modules remain entirely inactive (Tab. \ref{tab:efficiency}) since \ourmodel{} has fully internalized the spatial grounding expertise during the two-stage training. Specifically, the triggering of the \lvrbox{lvr\_start} token initiates the latent reasoning driven purely by the autoregressive latent decoding process, where the hidden states act as dynamic probes to extract evidence from the global visual scene without relying on any external modules.

    \vspace{-10pt}
\section{Experiments}
\label{sec:Experiments}
\subsection{Experiment Setup}
\noindent \textbf{Implementation Details.} We adopt Qwen2.5-VL-7B-Instruct as the backbone, keeping the visual encoder and multimodal projector frozen while updating only the LLM parameters. The image is set to a resolution range of $128$ to $5120 \times 28 \times 28$ pixels. We utilize Visual CoT \cite{shao2024visual} dataset and handle variable sequence lengths via an adaptive multimodal data-packing strategy \cite{chen2024expanding}. Specifically, we limit each batch to 4 instances and apply a 4,096-token threshold for long sequences to guide dynamic grouping, resulting in a maximum packed sequence length of 16,384 tokens. Training is performed using BF16 precision with gradient checkpointing and DeepSpeed ZeRO-3 on an 8$\times$ NVIDIA A800 GPU cluster. We employ the AdamW optimizer with a 0.1 weight decay, a 3\% warmup phase, and a cosine learning rate schedule peaking at $1 \times 10^{-5}$ for Stage 1 and $5 \times 10^{-6}$ for Stage 2, requiring around 24/12 hours for 2,500/1,250 steps for each stage. $\lambda_{BCM}$, $\lambda_{DAC}$, $\lambda_{feat}$ and $\lambda_{attn}$ are set to 0.1, 0.1, 1, and 1, respectively. The latent reasoning steps $S$ are set to 8 during both training and inference. 

\noindent \textbf{Evaluated Benchmarks.} We evaluate \ourmodel{} across a diverse set of vision-centric benchmarks focusing on fine-grained perception and high-resolution understanding. We utilize MMVP \cite{tong2024eyes} to measure perception robustness and V$^*$ Bench \cite{wu2024v} for detailed visual search. For higher-level cognitive tasks, we adopt the BLINK \cite{fu2024blink} benchmark, covering tasks such as Counting, IQ-Test, JigSaw, Relative Reflectance, and Spatial Relation. We also evaluate \ourmodel{} on high-resolution benchmarks, including HRBench-4K \cite{hrbench}, HRBench-8K \cite{hrbench}, and MME-Real-Lite \cite{zhang2024mme}. Specifically, within the HR-Bench suites, Fine-grained Single-instance Perception (FSP) evaluates the recognition of individual object attributes, and Fine-grained Cross-instance Perception (FCP) assesses the model's ability to reason over complex relationships and global layouts across multiple instances. These benchmarks collectively quantify the model's ability to process and reason over real-world, high-pixel visual evidence.

\subsection{Main Results}

\begin{table*}[t]
\centering
\caption{Performance comparison on visual perception and cognitive benchmarks. Comparison results are reported by \cite{li2025latent}.}
\resizebox{\textwidth}{!}{
    \setlength{\tabcolsep}{3pt}
    \begin{tabular}{l c | c | cccccc | ccc}
    \toprule
    \multirow{3}{*}{\textbf{Method}} & \multirow{3}{*}{\textbf{\makecell[c]{Param \\ Size}}} & \multicolumn{1}{c|}{\textbf{MMVP}} & \multicolumn{6}{c|}{\textbf{BLINK}} & \multicolumn{3}{c}{\textbf{V*}} \\
    \cmidrule(lr){3-3} \cmidrule(lr){4-9} \cmidrule(lr){10-12}
    
    & & \textbf{Overall} & \textbf{Overall} & \textbf{Counting} & \textbf{IQ-Test} & \textbf{JigSaw} & \makecell[c]{\textbf{Relative}\\\textbf{Reflect}} & \makecell[c]{\textbf{Spatial}\\\textbf{Relation}} & \textbf{Overall} & \textbf{V$^*_{D.A.}$} & \textbf{V$^*_{R.P.}$} \\ \midrule
    \rowcolor{orange!5} \multicolumn{12}{c}{\textit{Proprietary Model}} \\
    GPT-4o \cite{hurst2024gpt} & - & 58.33 & 51.1 & 51.7 & 30.0 & 58.0 & 38.8 & 76.9 & 62.8 & - & - \\
    \midrule
    \rowcolor{blue!5} \multicolumn{12}{c}{\textit{Open-Source Model}} \\
    Qwen2.5-VL-7B & 7B & 66.7 & 54.5 & 65.8 & 27.3 & 52.7 & 41.0 & 88.1 & 78.5 & 81.7 & 73.7 \\
    \quad + vanilla SFT & 7B & 69.5 & 53.1 & 60.8 & 26.7 & 45.3 & 33.6 & 88.8 & 79.1 & 82.6 & 73.7 \\
    PAPO \cite{wang2025perception} & 7B & 54.3 & 54.8 & 66.7 & 29.3 & 52.0 & 39.6 & 88.8 & 36.1 & 25.2 & 52.6 \\
    Vision-R1 \cite{huang2025vision} & 7B & 46.7 & 42.8 & 51.7 & 26.7 & 27.3 & 44.8 & 66.4 & 70.2 & 70.4 & 69.7 \\
    PixelReasoner \cite{wang2025pixel} & 7B & 67.0 & 54.5 & 66.7 & 25.3 & 52.7 & 42.5 & 88.1 & 80.1 & 81.7 & 77.6 \\
    LVR \cite{li2025latent} & 7B & 71.7 & 55.4 & 70.0 & 29.3 & 52.0 & 42.5 & 86.0 & 81.7 & 84.4 & 77.6 \\
    \midrule
\rowcolor{green!5} \multicolumn{12}{c}{\textit{Our Model}} \\
    \ourmodel{} (ours) & 7B & \textbf{72.3} & \textbf{56.4} & 65.8 & \textbf{33.3} & 49.6 & \textbf{44.8} & \textbf{90.9} & \textbf{81.7} & 83.5 & \textbf{78.9} \\ 
    \rowcolor{green!5} \textbf{$\Delta$ (vs. Qwen2.5-VL-7B)} & - & \textbf{+5.6} & \textbf{+1.9} & 0.0 & \textbf{+6.0} & -3.1 & \textbf{+3.8} & \textbf{+2.8} & \textbf{+3.2} & \textbf{+1.8} & \textbf{+5.2} \\ \bottomrule
    \end{tabular}
}
\label{table:bench_table1}
\end{table*}

\begin{table*}[t]
\centering
\small
\caption{Performance on real-world high-resolution perception and reasoning benchmarks. Results marked with $^{\dagger}$ are reported by \cite{zhang2025thyme}, while others are reported by \cite{wang2025monet}.}
\resizebox{\textwidth}{!}{
\begin{tabular}{lc | ccc | ccc | ccc}
\toprule
\multirow{2}{*}{\textbf{Method}} & \multirow{2}{*}{\textbf{\makecell[c]{Param \\ Size}}} & \multicolumn{3}{c|}{\textbf{HRBench-4K}} & \multicolumn{3}{c|}{\textbf{HRBench-8K}} & \multicolumn{3}{c}{\textbf{MME-Real-Lite}} \\
\cmidrule(lr){3-5} \cmidrule(lr){6-8} \cmidrule(lr){9-11}
 & & Overall & FSP & FCP & Overall & FSP & FCP & Overall & Reasoning & Perception \\
\midrule
\rowcolor{orange!5} \multicolumn{11}{c}{\textit{Proprietary Model}} \\
GPT-4o \cite{hurst2024gpt} & - & 65.0 & 66.8 & 63.3 & 59.6 & 60.8 & 58.5 & 52.0 & 48.3 & 54.4 \\
\midrule
\rowcolor{blue!5} \multicolumn{11}{c}{\textit{Open-Source Model}} \\
Qwen2.5-VL-7B & 7B & 68.0 & 80.3 & 55.8 & 63.8 & 73.8 & 51.8 & 45.8 & 39.7 & 49.6 \\
\quad + vanilla SFT & 7B & 69.2 & 78.8 & 59.8 & 64.5 & 76.8 & 52.3 & 50.1 & 42.5 & 54.1 \\
InternVL3 \cite{zhu2025internvl3}& 8B & 70.0$^{\dagger}$ & 78.8$^{\dagger}$ & 61.3$^{\dagger}$ & 69.3$^{\dagger}$ & 78.8$^{\dagger}$ & 59.8$^{\dagger}$ & 48.6$^{\dagger}$ & 44.8$^{\dagger}$ & 51.0$^{\dagger}$ \\
Deepeyes \cite{zheng2025deepeyes} & - & 71.3 & 83.8 & 58.8 & 65.1 & 77.0 & 53.3 & 54.3 & 50.5 & 56.6 \\
\midrule
\rowcolor{green!5} \multicolumn{11}{c}{\textit{Our Model}} \\
\ourmodel{} & 8B & \textbf{72.6} & \textbf{83.5} & \textbf{61.8} & 66.3 & 73.5 & \textbf{58.5} & 53.9 & 45.0 & \textbf{58.5} \\
\rowcolor{green!5} \textbf{$\Delta$ (vs. Qwen2.5-VL-7B)} & - & \textbf{+4.6} & \textbf{+3.2} & \textbf{+6.0} & \textbf{+2.5} & \textbf{-0.3} & \textbf{+6.7} & \textbf{+8.1} & \textbf{+5.3} & \textbf{+8.9} \\
\bottomrule
\end{tabular}
}
\label{table:bench_table2}
\end{table*}

As shown in Tab. \ref{table:bench_table1} and Tab. \ref{table:bench_table2}, \ourmodel{} demonstrates substantial performance gains over state-of-the-art 7B/8B scale models across a diverse set of benchmarks: (1) \textbf{Robustness in Perception and Cognition}: On MMVP, \ourmodel{} achieves a score of 72.3, significantly outperforming Qwen2.5-VL-7B by 7.0 points and even surpassing GPT-4o (58.33). This highlights its ability to overcome common CLIP-blind limitations. Within the BLINK cognitive suite, our model shows a +1.4 overall improvement, particularly excelling in tasks requiring complex mental manipulation. Furthermore, it achieves a competitive 81.7 on the V$^*$ Bench, demonstrating its efficacy in high-precision visual search. (2) \textbf{Superiority in High-Resolution Reasoning}: On HRBench-4K/8K, our model obtains overall scores of 72.6 and 66.3, respectively, maintaining a clear lead over InternVL3 and Deepeyes. Most notably, \ourmodel{} exhibits a +6.7 leap in FCP on HRBench-4K, underscoring its advanced capacity for reasoning over complex spatial layouts and multi-object relationships in high-fidelity images. (3) \textbf{Real-World Benchmarking}: On the challenging MME-Real-Lite benchmark, \ourmodel{} achieves an overall accuracy of 53.9. While maintaining strong logical reasoning, its perception score of 58.5 represents a dominant +8.9 increase over the baseline, validating the model's superior ability to ground its outputs in authentic, high-pixel visual evidence.

\subsection{Ablation Studies}

\begin{table*}[t]
\centering
\caption{\textbf{Ablation Studies of \ourmodel{}.} We evaluate \ourmodel{} across four dimensions: (a) contribution of core modules BCM and DAC; (b) effectiveness of $\mathcal{L}_{BCM}$ \& $\mathcal{L}_{DAC}$ Loss; (c) necessity of the two-stage training paradigm; (d) impact of different $Q_{dyn}$ sources for the DAC module; (e) inference efficiency analysis; and (f) impact of latent steps ($S$).}
\label{tab:comprehensive_ablation}
\vspace{5pt}
\begin{minipage}{0.43\textwidth}
\centering
\text{(a) BCM \& DAC Module Ablation} \\
\resizebox{\textwidth}{!}{
    \begin{tabular}{cc | ccc}
    \toprule
    \textbf{BCM} & \textbf{DAC} & \textbf{MMVP} & \textbf{HRBench-4K} & \textbf{BLINK} \\
    \midrule
    - & - & 69.5 & 68.0 & 53.1 \\
    \checkmark & - & 70.9 & 71.1 & 54.2 \\
    - & \checkmark & 65.9 & 66.8 & 52.2 \\
    \rowcolor{green!5} \checkmark & \checkmark & \textbf{72.3} & \textbf{72.6} & \textbf{56.4} \\
    \bottomrule
    \end{tabular}
}
\end{minipage}
\hfill
\begin{minipage}{0.48\textwidth}
\centering
\text{(b) $\mathcal{L}_{BCM}$ \& $\mathcal{L}_{DAC}$ Loss Ablation} \\
\resizebox{\textwidth}{!}{
    \begin{tabular}{l | ccc}
    \toprule
    \textbf{Strategy} & \textbf{MMVP} & \textbf{HRBench-4K} & \textbf{BLINK} \\
    \midrule
    w/o sg(.) ($\mathcal{L}_{BCM}$) & 68.4 & 67.1 & 52.6 \\
    $\lambda_{\text{feat}}=0$ ($\mathcal{L}_{DAC}$) & 71.5 & 71.2 & 54.4 \\
    $\lambda_{\text{attn}}=0$  ($\mathcal{L}_{DAC}$) & 70.9 & 70.1 & 53.6 \\
    \rowcolor{green!5} \textbf{Full Framework} & \textbf{72.3} & \textbf{72.6} & \textbf{56.4} \\
    \bottomrule
    \end{tabular}
}
\end{minipage}

\begin{minipage}{0.48\textwidth}
\centering
\vspace{5pt}
\text{(c) Training Stage Ablation} \\
\resizebox{\textwidth}{!}{
    \begin{tabular}{l | ccc}
    \toprule
    \textbf{Training Setup} & \textbf{MMVP} & \textbf{HRBench-4K} & \textbf{BLINK} \\
    \midrule
    Stage 1 Only & 70.9 & 71.1 & 54.2 \\
    Stage 2 Only & 68.4 & 67.5 & 52.9 \\
    Joint Train & 70.2 & 70.1 & 53.6 \\
    \rowcolor{green!5} \textbf{Two-stage (Ours)} & \textbf{72.3} & \textbf{72.6} & \textbf{56.4} \\
    \bottomrule
    \end{tabular}
}
\end{minipage}
\hfill
\begin{minipage}{0.48\textwidth}
\centering
\text{(d) $Q_{dyn}$ in DAC Module} \\
\resizebox{\textwidth}{!}{
    \begin{tabular}{l | ccc}
    \toprule
    \textbf{Source} & \textbf{MMVP} & \textbf{HRBench-4K} & \textbf{BLINK} \\
    \midrule
    Random Gaussian & 37.3 & 35.0 & 28.2 \\
    Static Learned $Q_{dyn}$ & 67.5 & 66.1 & 52.8 \\
    \rowcolor{green!5} \textbf{Latent States $\mathbf{H}$} & \textbf{72.3} & \textbf{72.6} & \textbf{56.4} \\
    \bottomrule
    \end{tabular}
}
\end{minipage}

\begin{minipage}[t]{0.66\textwidth}
    \centering
    \vspace{5pt}
    \text{(e) Inference Efficiency Analysis.} \\
    \label{tab:efficiency}
    \resizebox{\linewidth}{!}{
    \begin{tabular}{lcc}
        \toprule
        \textbf{Metric} & \textbf{Baseline (7B)} & \textbf{\ourmodel{}} \\
        \midrule
        \textbf{Active Resamplers} & None & None \\
        \textbf{Params Overhead} & 0 & 102.83M (+1.45\%) \\
        \textbf{VRAM (bf16)} & 0 & 196 MB (+1.5\%) \\
        \textbf{Extra Arch. FLOPs} & 0 & 0 \\
        \textbf{Reasoning Cost} & 0 steps & +$S$ Latent Steps \\
        \textbf{Mechanism} & Standard & Coconut-style\cite{hao2024training} \\
        \bottomrule
    \end{tabular}
    }
\end{minipage}
\hfill
\begin{minipage}[t]{0.30\textwidth}
    \vspace{3pt} 
    \centering
    \includegraphics[width=\linewidth]{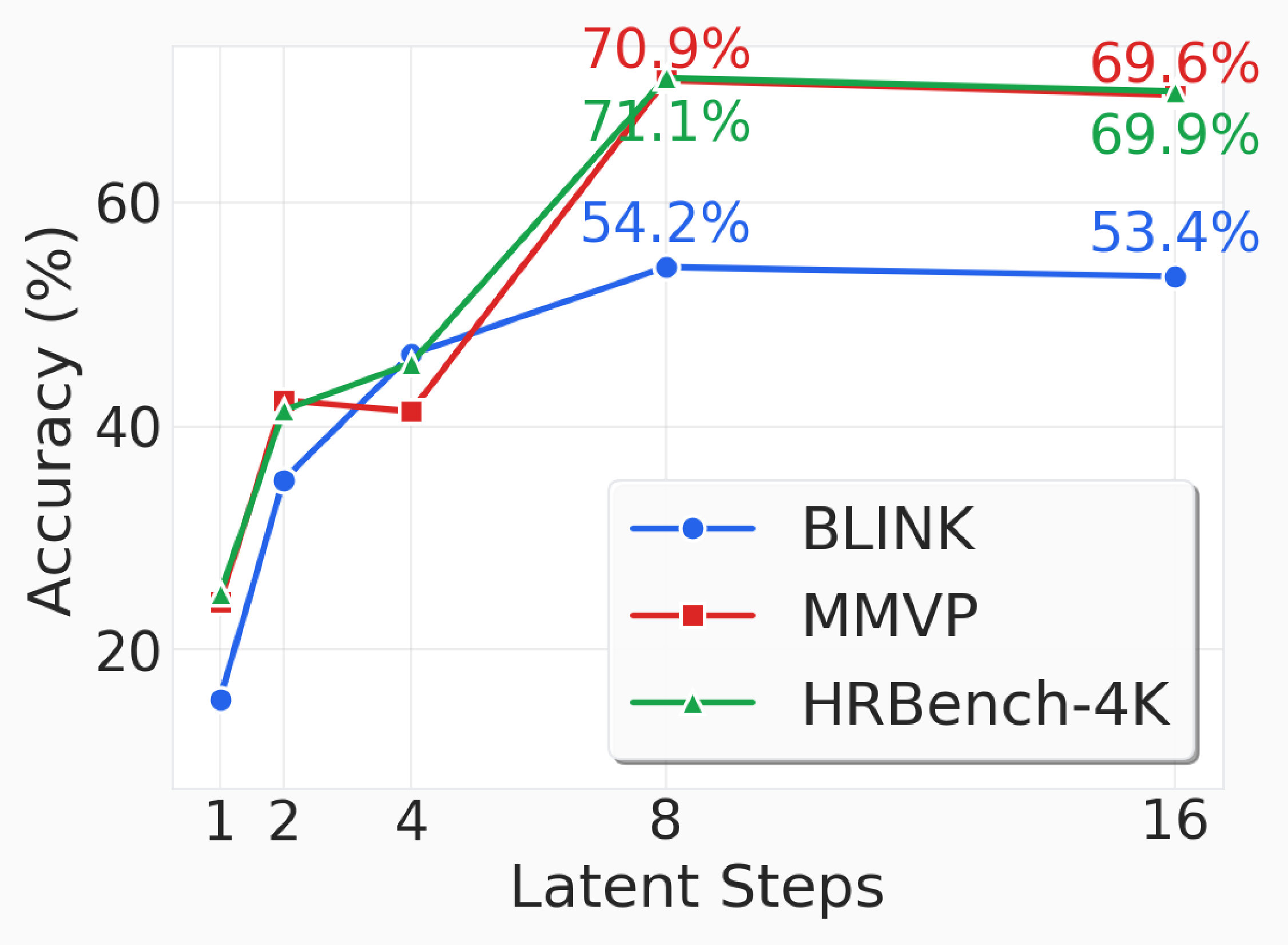}
    \text{(f) Impact of Latent Steps ($S$).}
\end{minipage}

\label{tab:ablation}
\end{table*}

\noindent \textbf{Effectiveness of BCM and DAC Modules}: As shown in Tab.~\ref{tab:ablation} (a), removing either module leads to a consistent performance drop. The vanilla model (without BCM and DAC) only achieves 69.5 on MMVP and 68.0 on HRBench-4K. While BCM alone provides a solid baseline for grounded perception, its synergy with DAC yields the most significant gains (+2.8 on MMVP over BCM alone). This confirms that the transition from explicit box-guided perception to implicit latent reasoning is crucial for the model to internalize grounding capabilities.

\noindent \textbf{Impact of Loss Functions}: Tab.~\ref{tab:ablation} (b) highlights the necessity of the stochastic decoupled alignment strategy ($\mathcal{L}_{BCM}$) and dual-objective distillation ($\mathcal{L}_{DAC}$). Removing the stop-gradient operator ($\text{sg}(\cdot)$) in $\mathcal{L}_{BCM}$ leads to severe degradation (68.4 on MMVP), proving our hypothesis of representation collapse during joint optimization. Furthermore, removing either the feature distillation ($\lambda_{\text{feat}}=0$) or the attention distillation ($\lambda_{\text{attn}}=0$) leads to suboptimal performance, proving that both semantic and spatial guidance are vital for the DAC module.

\noindent \textbf{Two-stage Training Paradigm}: Tab.~\ref{tab:ablation} (c), our two-stage paradigm is vital for transitioning from explicit perception to autonomous reasoning. Training only on Stage 2 (68.4 on MMVP) causes the model to struggle with localizing key evidence without the supervision of grounded spatial features from Stage 1. Additionally, the \textit{Joint Train} approach (70.2) underperforms the two-stage strategy, likely due to the convergence challenge when optimizing teacher and student modules simultaneously. By sequentially establishing latent-to-pixel alignment before distilling it into the DAC module, we enable a stable transition to targeted visual probing from global context.

\noindent \textbf{Dynamic Query $Q_{dyn}$ Source}: In Tab.~\ref{tab:ablation} (d), we evaluate various $Q_{dyn}$ queries to identify the optimal mechanism for visual latent reasoning. Utilizing the linear projection $\phi(.)$ of LLM's latent states $\mathbf{H}$ as dynamic queries achieves superior performance, reaching 72.3 on MMVP and 72.6 on HRBench-4K. Unlike static learned queries (67.5 on MMVP) that rely on fixed visual bottlenecks, our dynamic approach enables the model to adaptively interrogate the scene based on its reasoning process. The failure of random Gaussian probes (37.3 on MMVP) further underscores DAC's effectiveness stems from the semantic alignment between evolving thoughts and targeted visual evidence. 

\noindent \textbf{Latent reasoning steps $S$}:
We investigate the impact of latent reasoning steps S on model performance during training stage 1. As illustrated in Fig. \ref{tab:ablation} (f), accuracy across all benchmarks significantly improves as S increases from 1 to 8, with MMVP performance peaking at 70.9\%. This trend suggests that sufficient latent iterations are essential for the reasoning trajectory to effectively probe and resolve complex visual ambiguities. Performance plateaus or slightly declines beyond S=8, dropping to 69.6\% on MMVP at S=16. Consequently, we select S=8 as our default configuration to achieve an optimal balance between peak perception accuracy and inference efficiency.

\subsection{Efficiency Analysis}

As shown in Tab.~\ref{tab:ablation} (e), \ourmodel{} demonstrates exceptional inference efficiency by keeping its two training distillation modules (BCM \& DAC modules) entirely inactive during inference. This introduces a minimal parameter overhead of just 1.45\% (196 MB) for a 7B base model, resulting in zero architectural FLOPs overhead during the forward pass. The sole computational cost stems from the Coconut-style \cite{hao2024training} continuous latent reasoning mechanism, which directly routes the last hidden state as the subsequent input embedding. This process adds exactly $S$  KV-cached decoding steps per reasoning cycle. Because these autoregressive steps are highly optimized, they incur only a manageable 25\% to 80\% latency increase for standard 10-to-30-token tasks. Ultimately, this design enables \ourmodel{} achieving visual self-reflection purely through its hidden states, preserving a highly efficient, pure-LLM decoding pathway.

\begin{figure}[tb]
  \centering
  \includegraphics[height=6cm]{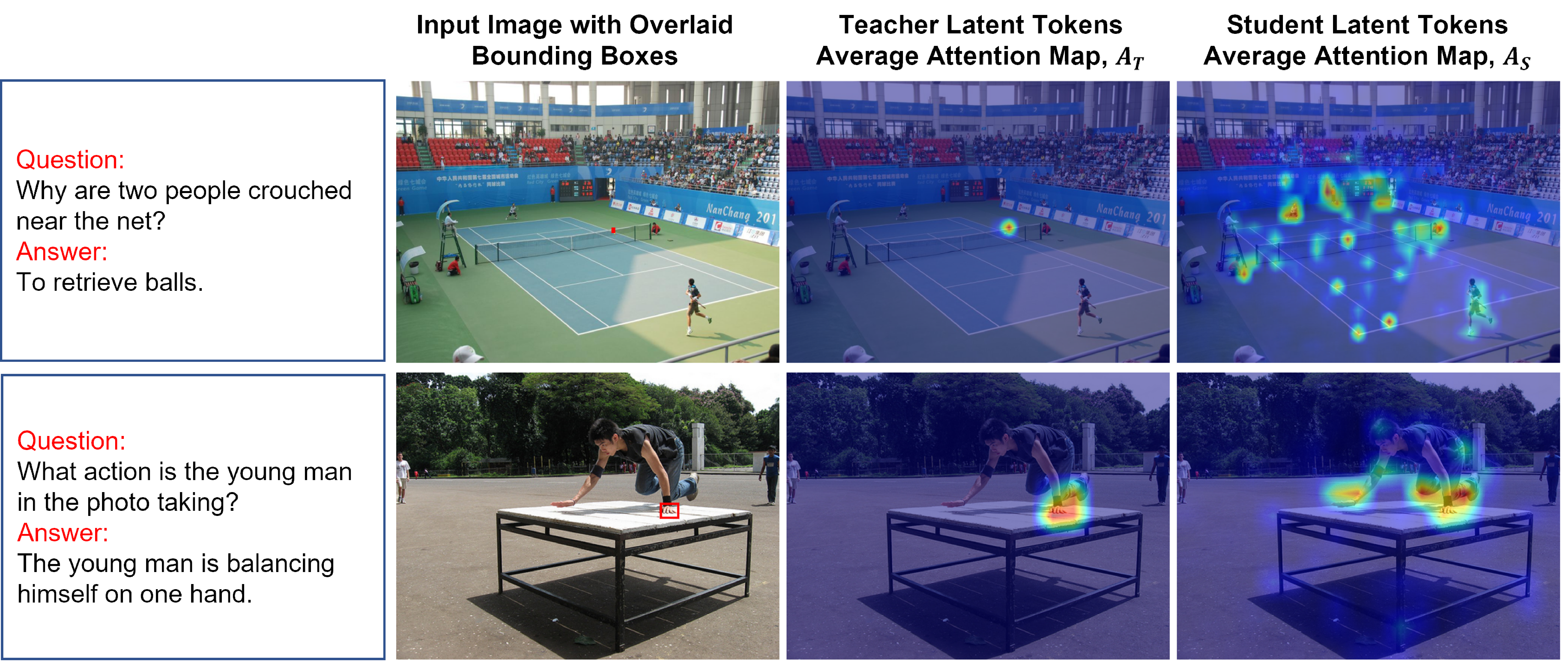}
\vspace{-10pt}
  \caption{\textbf{Visualization of Latent Reasoning during Training.} Averaged attention maps across all reasoning steps demonstrate that while the teacher (BCM) is confined to local priors, the student (DAC) successfully transcends bounding box constraints to capture global contextual relationships.}
  \label{fig3-train-atten}
  \vspace{-10pt}
\end{figure}

\subsection{Visualization Analysis: How Latent Tokens Help Visual Self-Reflection Reasoning?}
\noindent \textbf{Visualization of Visual Latent Distillation (Train).} 
Fig. \ref{fig3-train-atten} visualizes the average attention maps between
teacher/student latent tokens and image features across all reasoning steps. While the teacher’s attention is strictly localized within ground-truth bounding boxes with missing broader scene context, the student model autonomously explores global features through its evolving latent reasoning states. In Row 1, while the teacher's attention is restricted to the tennis ball in the middle of the picture, the student model captures the full spatial interaction between the crouched individuals and active players to infer their intent. Similarly, in Row 2, the student expands from the teacher’s local hand-level focus to a global body-pose analysis, enabling precise action recognition. These cases demonstrate the student's ability to transcend bounding-box constraints through latent reasoning. (\textit{See more visualization results in Suppl.})

\begin{figure}[tb]
  \centering
  \includegraphics[height=16cm]{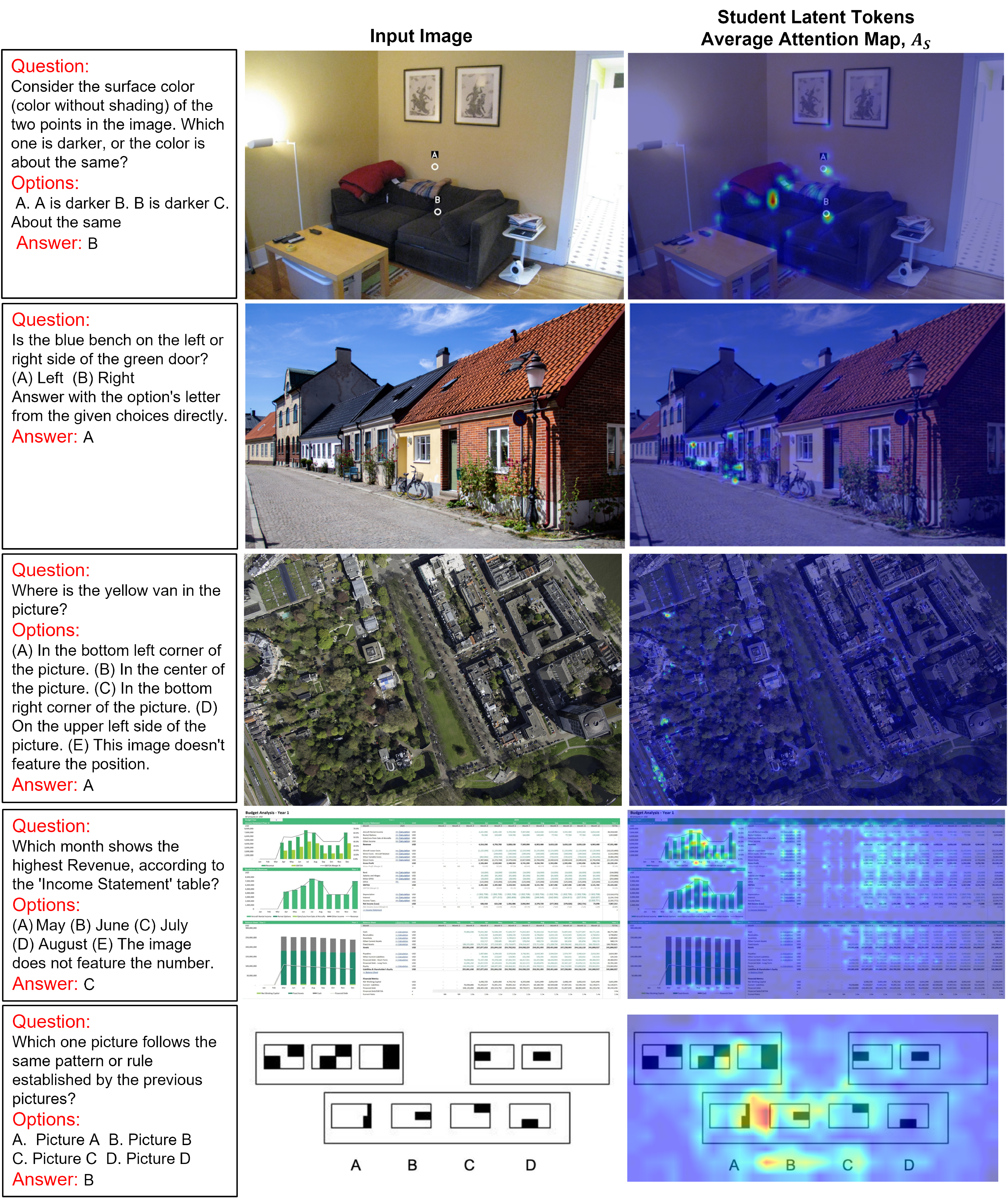}
  \caption{\textbf{Visualization of Latent Reasoning during Inference.} Averaged attention maps across all latent reasoning steps (Col. 3) autonomously pinpointing visual evidence driven by latent states.}
  \label{fig4-inference-atten}
\end{figure}

\noindent \textbf{Visualization of Visual Latent Reasoning (Inference).} Fig. \ref{fig4-inference-atten} visualizes the average attention map between student latent tokens and image features across all reasoning steps during the inference phase, where explicit bounding-box priors are absent. Guided by evolving latent states, V-Reflection dynamically localizes task-relevant visual evidence from global features. For instance, in Row 1, it demonstrates fine-grained precision by pinpointing specific coordinates (Points A and B) for surface-color queries. In Row 2, the model isolates the blue bench from the background to extract spatial information. These cases collectively highlight the model’s visual self-reflection mechanism, which enables autonomous, targeted probing for both object-level localization and pixel-level reasoning across diverse scenarios. (\textit{See more visualization results in Suppl.})

\section{Conclusion}
In this paper, we introduced \ourmodel{}, a novel framework that instantiates a "think-then-look" visual reflection mechanism. Through a two-stage distillation process, \ourmodel{} transfers explicit spatial grounding expertise from a Box-Guided Compression (BCM) teacher to a Dynamic Autoregressive Compression (DAC) student. This allows the model to internalize the ability to autonomously interrogate the visual context using its internal latent states. Experiment results across multiple perception-intensive tasks demonstrate that \ourmodel{} effectively mitigates fine-grained hallucinations while maintaining optimal inference efficiency with zero additional architecture. Visualizations further confirm the model’s ability to direct spatial attention toward task-critical evidence, marking a significant step toward truly grounded multimodal reasoning.

\noindent \textbf{Future Work} We plan to extend this active paradigm to dynamic modalities like video and embodied AI. Additionally, future iterations will explore reinforcement learning with verifiable rewards and adaptive probing to trigger visual searches based on uncertainty.

\newpage
{
    \small
    \bibliographystyle{plain}
    \bibliography{main}
}

\appendix
\appendix
\section{Architecture Specifications and Implementation Details}

To facilitate reproducibility, we provide detailed architectural specifications for the core modules introduced in our framework: the Box-Guided Compression Module (BCM) and the Dynamic Autoregressive Compression (DAC). 

\subsection{Architecture Specifications }
Both the BCM (Teacher) and DAC (Student) modules are instantiated as lightweight cross-attention mechanisms designed to bridge the visual encoder and the LLM's latent space. To minimize computational overhead and parameter count, both modules share an identical, minimalist architectural configuration:
\begin{itemize}
    \item \textbf{Number of Layers:} 1
    \item \textbf{Number of Attention Heads:} 8
    \item \textbf{Hidden Dimension ($D$):} Aligned with the LLM's hidden size. For our Qwen2.5-VL-7B backbone, $D = 3584$.
\end{itemize}

\noindent \textbf{Box-Guided Compression Module (BCM):} Operating exclusively during Stage 1 training, the BCM functions as the explicit spatial teacher. It employs a set of static, learnable queries $\mathbf{Z}_T \in \mathbb{R}^{S \times D}$ to compress regional visual features into latent representations. The sequence length of these learnable queries is strictly set to $S = 8$, corresponding to the fixed number of latent slots in the reasoning stream. Notably, bounding boxes are not injected as textual prompts; instead, they are provided as normalized coordinates ($[x_0, y_0, x_1, y_1] \in [0, 1]$) and converted into image token indices to precisely pool the regional visual features for the BCM's cross-attention.

\noindent \textbf{Dynamic Autoregressive Compression (DAC):} Introduced in Stage 2, the DAC acts as the student module. Rather than relying on static queries and localized bounding boxes, the DAC utilizes the LLM's dynamically evolving autoregressive hidden states as queries to attend to the global visual feature map. This 1-layer cross-attention produces the $S=8$ latent tokens that are subsequently aligned with the BCM's targets.

\subsection{Tokenization and Reasoning Formatting}
To accommodate the continuous latent reasoning within the standard LLM vocabulary, we expand the tokenizer with a set of dedicated special tokens:
\begin{itemize}
    \item \texttt{<|sovt|>}: A control token signaling the initiation of the continuous hidden-state reasoning mode.
    \item \texttt{<|lvr|>}: Represents the actual latent slots. Exactly 8 instances of this token are appended to accommodate the features outputted by the BCM or DAC.
    \item \texttt{<|eovt|>}: A control token marking the termination of the latent reasoning interval.
\end{itemize}
During Stage 1 training, the \texttt{<|lvr|>} token embeddings are directly replaced by the outputs of the BCM. In Stage 2 and inference, the DAC populates these slots dynamically.

\begin{algorithm}[htbp]
\caption{Two-Stage Training Algorithm}
\label{alg:training}
\begin{algorithmic}[1] 
\Require LLM hidden states $\mathbf{H}$, Learnable queries $\mathbf{Q}$, Global features $\mathbf{F}_{global}$, Grounding box $\mathcal{B}$.
\Require ROIAlign operator, BCM Module, DAC Module, LLM Backbone.
\Require Loss weights $\lambda_{BCM}, \lambda_{DAC}, \lambda_{feat}, \lambda_{attn}$, Temperature $\tau$, Smoothing $\epsilon$.
\Statex
\Statex \textbf{\textit{Stage 1: Explicit Grounding Warm-up. (Train LLM and BCM)}}
\State $\mathbf{F}_{local} \gets \text{ROIAlign}(\mathbf{F}_{global}, \mathcal{B})$ \Comment{Extract local features from global map}
\State $\mathbf{Z}_{T}, \mathcal{A}_T \gets \text{BCM}(\text{Query}=\mathbf{Q}, \text{Key/Value}=\mathbf{F}_{local})$ 
\State $\mathcal{L}_{CE} \gets \text{CrossEntropy}(\text{LLM}(\mathbf{Z}_{T}), \text{Target Labels})$
\State $\hat{\mathbf{H}} \gets \text{clamp}(\mathbf{H}, -10^4, 10^4)$ \Comment{Numerical stability}
\State $\hat{\mathbf{Z}}_{T} \gets \text{clamp}(\mathbf{Z}_{T}, -10^4, 10^4)$
\State Sample $\mathbb{I} \sim \text{Bernoulli}(0.5)$
\If{$\mathbb{I} = 1$}
    \State $\mathcal{L}_{BCM} \gets \|\text{sg}(\hat{\mathbf{Z}}_{T}) - \hat{\mathbf{H}}\|_1$ \Comment{Align BCM to LLM}
\Else
    \State $\mathcal{L}_{BCM} \gets \|\hat{\mathbf{H}} - \text{sg}(\hat{\mathbf{Z}}_{T})\|_1$ \Comment{Align LLM to BCM}
\EndIf
\State $\mathcal{L}_{stage1} \gets \mathcal{L}_{CE} + \lambda_{BCM} \cdot \mathcal{L}_{BCM}$
\State Update BCM and LLM using $\nabla \mathcal{L}_{stage1}$

\Statex
\Statex \textbf{\textit{Stage 2: Visual Latent Distillation. (Train LLM and DAC)}}
\State Freeze BCM.
\State $\mathbf{F}_{local} \gets \text{ROIAlign}(\mathbf{F}_{global}, \mathcal{B})$ \Comment{Teacher still uses GT boxes for target generation}
\State $[\mathbf{Z}_{T}, \mathcal{A}_T] \gets \text{BCM}(\text{Query}=\mathbf{Q}, \text{Key/Value}=\mathbf{F}_{local})$ 
\State $\mathbf{Q}_{dyn} \gets \phi(\mathbf{H})$ \Comment{Dynamic probes from LLM states}
\State $[\mathbf{Z}_{S}, \mathcal{A}_S] \gets \text{DAC}(\text{Query}=\mathbf{Q}_{dyn}, \text{Key/Value}=\mathbf{F}_{global})$ \Comment{Uses global features}
\State $\mathcal{L}_{CE} \gets \text{CrossEntropy}(\text{LLM}(\mathbf{Z}_{S}), \text{Target Labels})$

\Statex \Comment{Visual Latent Distillation (VLD)}
\State Construct $\hat{\mathcal{A}}_T$ by projecting $\mathcal{A}_T$ onto global grid using $\mathcal{B}$.
\State $\mathcal{L}_{attn} \gets \sum_{i} D_{KL}(\hat{\mathcal{A}}_{T, i} \parallel \text{Softmax}(\mathcal{A}_{S, i} / \tau))$ 
\State $\mathcal{L}_{DAC} \gets \lambda_{feat} \cdot \|\mathbf{Z}_S - \text{sg}(\mathbf{Z}_T)\|_1 + \lambda_{attn} \cdot \mathcal{L}_{attn}$
\State $\mathcal{L}_{stage2} \gets \mathcal{L}_{CE} + \lambda_{DAC} \cdot \mathcal{L}_{DAC}$
\State Update DAC and LLM using $\nabla \mathcal{L}_{stage2}$
\end{algorithmic}
\end{algorithm}

\subsection{Algorithm for two-stage training framework}
Algorithm \ref{alg:training} details our two-stage training framework. In Stage 1 (Explicit Grounding Warm-up), the Box-Guided Compression Module (BCM) uses static queries $Q$ to extract target latent tokens $Z_{T}$ from localized features $\mathbf{F}_{local}$. To prevent representation collapse during alignment, we employ a stochastic decoupled alignment strategy: a Bernoulli indicator $\mathbb{I} \sim \text{Bernoulli}(0.5)$ directs the gradient flow via the stop-gradient operator $\text{sg}(\cdot)$, ensuring only one modality updates per iteration. In Stage 2 (Visual Latent Distillation), with the BCM frozen, the Dynamic Autoregressive Compression (DAC) learns to reconstruct the BCM's targets. It dynamically projects LLM hidden states into queries ($\mathbf{Q}_{dyn}$) to extract student latent tokens $Z_{S}$ from global image features $\mathbf{F}_{global}$. By minimizing the Mean Squared Error (MSE) between the DAC's predictions $Z_{S}$ and explicit spatial targets $Z_{T}$, localized visual expertise is effectively distilled into the model's autonomous, global reasoning stream.

\section{Ablation studies}

\noindent \textbf{Bernoulli probability $p$ in the Stochastic Decoupled Alignment Strategy.} As shown in Table \ref{tab:ablation_p}, we investigate the impact of $p$, the probability governing the Bernoulli distribution formulated in Eq. 2, within our stochastic decoupled alignment strategy. The results exhibit a clear inverted-U trend, with the model achieving its peak performance across all evaluated benchmarks (MMVP: 70.9, HRBench-4K: 71.1, and BLINK: 54.2) when $p$ is set to $0.5$.. Deviating from this balanced setting toward extreme values, such as $p=0.1$ or $p=0.9$, leads to a noticeable performance drop. This degradation suggests that an unbalanced probability overly biases the gradient updates toward either the LLM or the BCM teacher, thereby undermining the mutual feature alignment. Consequently, $p=0.5$ provides the necessary equilibrium for stable training, effectively maximizing the bidirectional grounding between the visual spatial queries and the latent reasoning manifold.

\begin{table}[htbp]
\centering
\caption{\textbf{Ablation Study on Bernoulli Probability $p$ in the Stochastic Decoupled Alignment Strategy.}}
\label{tab:ablation_p}
\vspace{-2mm}
\resizebox{0.5\linewidth}{!}{
\begin{tabular}{l c c c}
    \toprule
    $p$ & MMVP & HRBench-4K & BLINK \\
    \midrule
    0.1 & 69.5 & 69.1 & 53.1 \\
    0.3 & 69.8 & 70.4 & 53.7 \\
    \rowcolor{green!10} \textbf{0.5} & \textbf{70.9} & \textbf{71.1} & \textbf{54.2} \\
    0.7 & 70.4 & 70.2 & 53.5 \\
    0.9 & 70.1 & 69.7 & 52.9 \\
    \bottomrule
\end{tabular}}
\end{table}

\section{Additional Visualization Results}
\begin{figure*}[tb]
  \centering
  \includegraphics[height=21cm]{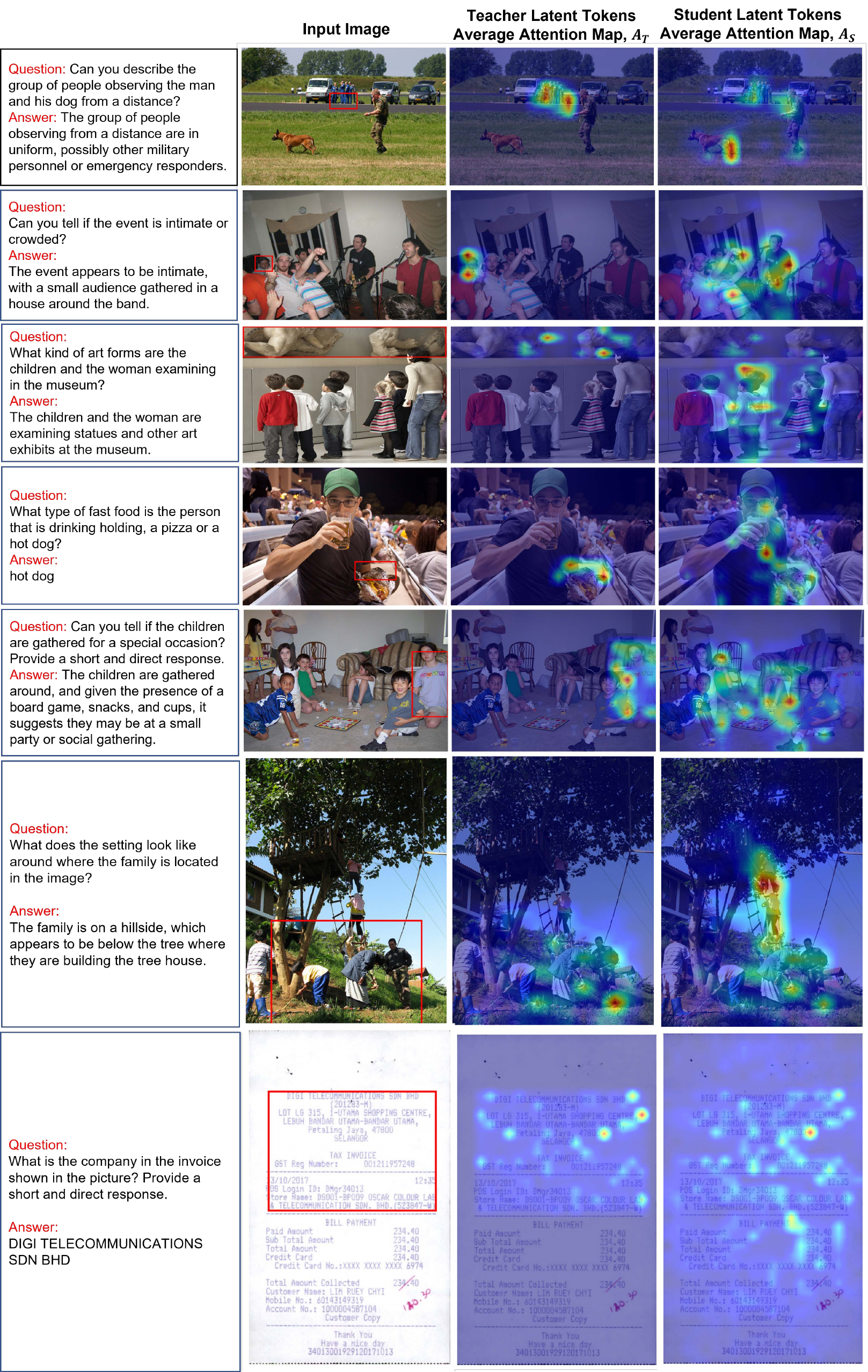}
  \caption{\textbf{Visualization of visual latent distillation during training.} While the teacher (BCM) is confined to local priors, the student (DAC) successfully transcends bounding box constraints to capture global contextual relationships.}
  \label{fig:supp_attn_train}
\end{figure*}

\noindent \textbf{Visualization of Visual Latent Distillation (Train).}
To provide further qualitative insights into the transition from localized grounding to autonomous visual interrogation, we present additional attention map visualizations for visual latent distillation in Fig. \ref{fig:supp_attn_train}. While the BCM teacher is explicitly tethered to local spatial priors (indicated by red bounding boxes in Col. 2), the DAC student consistently demonstrates the capability to transcend these constraints. As illustrated in the fifth row of Fig. \ref{fig:supp_attn_train}, \ourmodel{} exhibits a sophisticated ability to transcend the teacher's rigid spatial priors. While the BCM teacher is strictly confined to the red bounding box focused on the localized object, the DAC student autonomously extends its attention to encompass critical contextual elements, such as the subject’s face and the person-object interaction. This transition from passive regional observation to active, intent-driven interrogation confirms that the distillation paradigm successfully internalizes spatial expertise, empowering the model's internal reasoning manifold to pinpoint task-relevant evidence across the global visual field independently.

\noindent \textbf{Visualization of Visual Latent Reasoning (Inference).}
To further validate the generalization and robustness of our framework, we provide comprehensive visualizations of \ourmodel{}’s latent reasoning process across diverse, perception-intensive domains. During inference, the model utilizes its internal hidden states as dynamic probes to autonomously localize task-critical evidence without any external box priors or active distillation modules.

\begin{figure*}[tb]
  \centering
  \includegraphics[height=21cm]{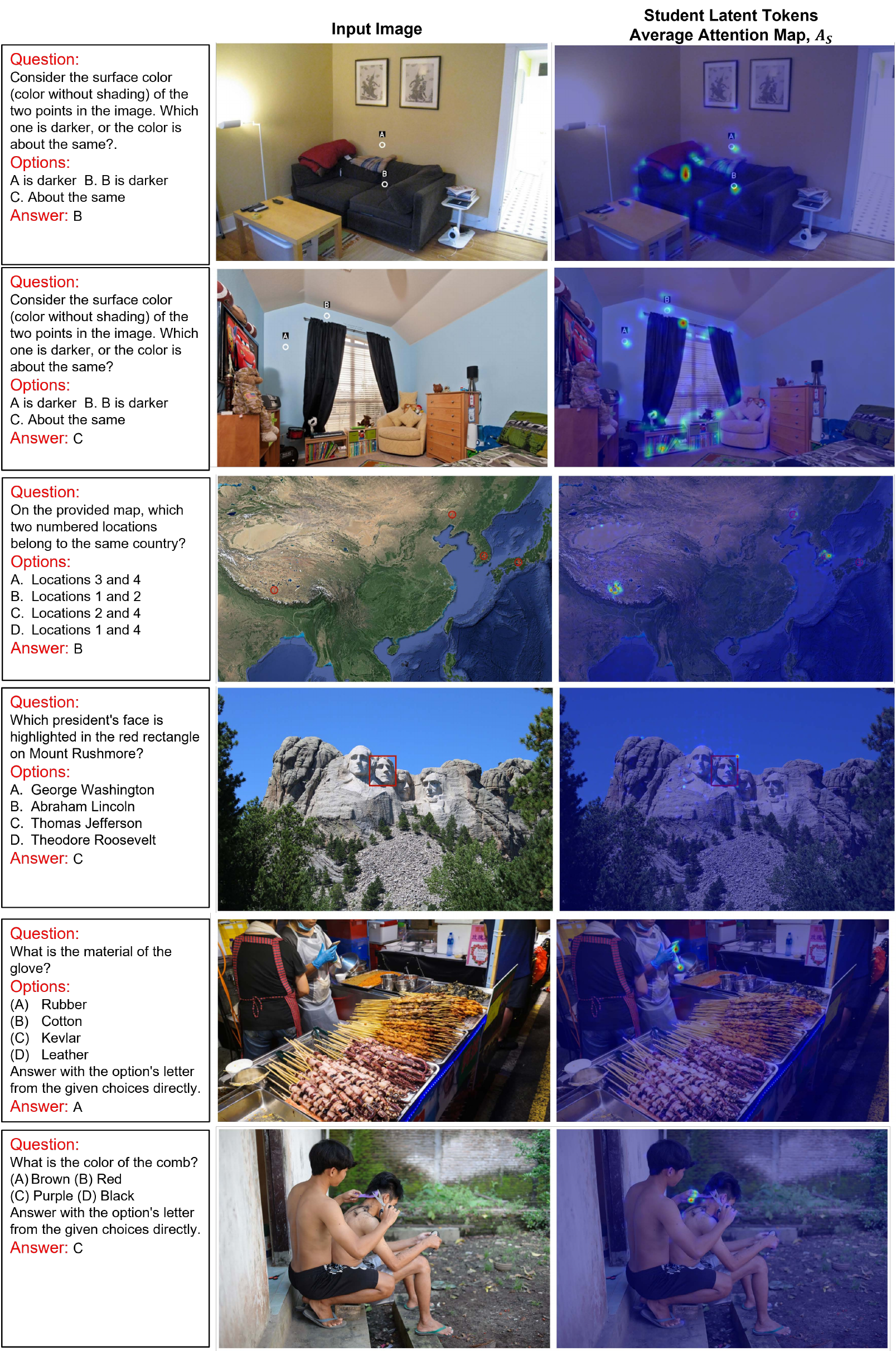}
  \caption{\textbf{Visualization of visual latent reasoning during inference.} Rows 1–4: cross-image; Rows 5–6: direct attribute.}
  \label{fig:supp_attn_infer1}
\end{figure*}

\textbf{Fine-Grained Attribute and Relational Reasoning:} As shown in Fig. \ref{fig:supp_attn_infer1}, the model excels at both cross-image comparative reasoning (Rows 1–4) and direct attribute identification (Rows 5–6). In the "Mount Rushmore" example (Row 4), the latent attention precisely targets the specific statue queried, while in the "street food" and "barber" scenes (Rows 5–6), it accurately pinpoints fine-grained interaction points between subjects and objects.

\begin{figure*}[tb]
  \centering
  \includegraphics[height=21cm]{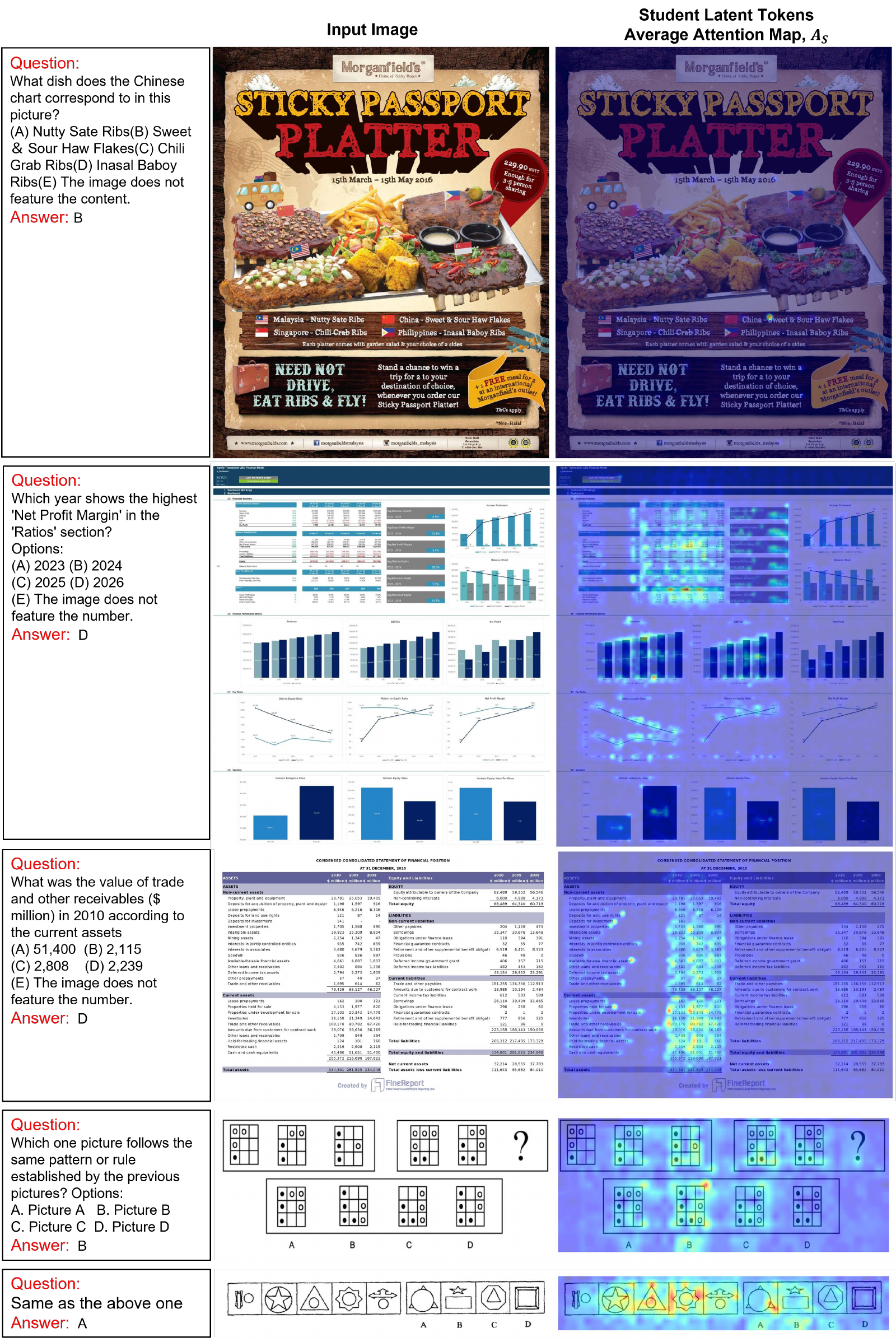}
  \caption{\textbf{Visualization of visual latent reasoning during inference.} Row 1: OCR; Rows 2–3: tables and diagrams; Rows 4–5: IQ tests.}
  \label{fig:supp_attn_infer2}
\end{figure*}

\textbf{Document and Abstract Logic Understanding:} Fig. \ref{fig:supp_attn_infer2} demonstrates the model's high-density information extraction capabilities. It effectively handles complex OCR tasks in menus (Row 1), navigates intricate tabular structures and trend lines in financial reports (Rows 2–3), and resolves abstract spatial logic in Raven’s Progressive Matrices and geometric IQ tests (Rows 4–5). The concentrated attention on specific cells or pattern components highlights a shift from global glancing to intent-driven evidence retrieval.

\begin{figure*}[tb]
  \centering
  \includegraphics[height=21cm]{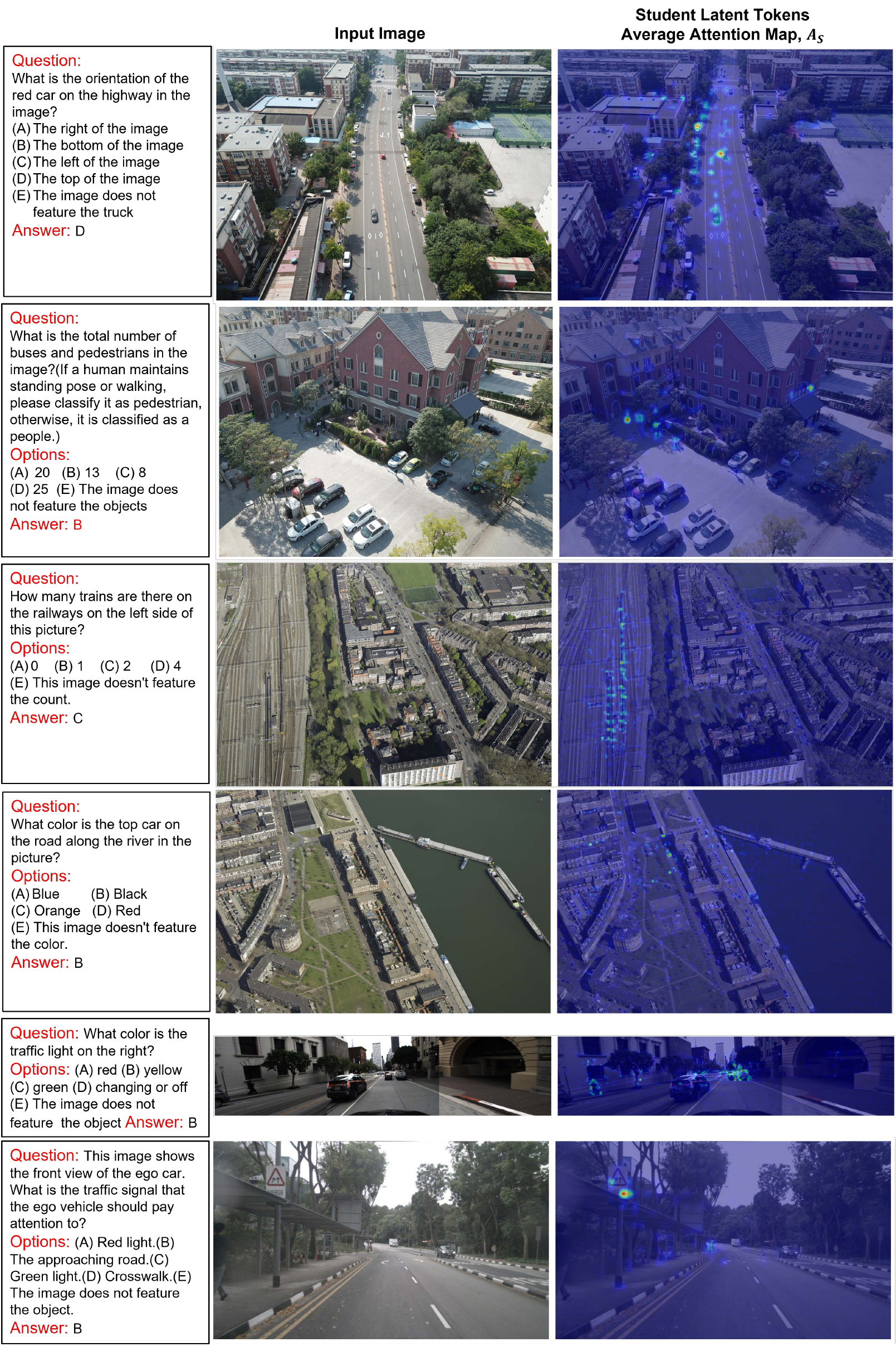}
  \caption{\textbf{Visualization of visual latent reasoning during inference.} Rows 1–2: monitoring; Rows 3–4: remote sensing; Rows 5-6: autonomous driving.}
  \label{fig:supp_attn_infer3}
\end{figure*}

\textbf{Specialized Domain Application:} The model's versatility is further evidenced in specialized scenarios in Fig. \ref{fig:supp_attn_infer3}. In monitoring and remote sensing (Rows 1–4), the probes successfully identify small-scale targets such as specific vehicles on a roadway or vessels in a harbor. Similarly, in autonomous driving contexts (Rows 5–6), the model focuses on critical safety cues, such as distant traffic signs and lane markers, proving its utility for high-stakes visual decision-making.

These diverse examples confirm that the internalized spatial expertise is not restricted to training categories but serves as a universal mechanism for pinpointing relevant visual evidence across a broad spectrum of real-world benchmarks.

\end{document}